\documentclass[11pt]{article}
\usepackage{acl2016}
\usepackage{times}
\usepackage{url}
\usepackage{latexsym}

\usepackage{amssymb}
\usepackage{amsthm}
\usepackage{color}
\usepackage{amsmath}
\usepackage{graphicx}
\usepackage{multirow}
\usepackage{breqn}
\usepackage{url}
\usepackage{comment}

\usepackage{subcaption}
\usepackage{caption}
\usepackage{tikz}
\usepackage{verbatim}

\usepackage{mathrsfs}
\usepackage{verbatim}
\usepackage{times}
\usepackage{url}
\usepackage{latexsym}
\usepackage{color}
\usepackage{float}
\usepackage{array}
\usepackage{amsmath,amsfonts,amssymb}
\usepackage{algorithm}
\usepackage{algpseudocode}
\usepackage{graphicx}
\usepackage{subcaption}
\usepackage{booktabs}
\usepackage{multirow}
\usepackage{tikz,pgfplots}
\usepackage{pgfplotstable}
\usepackage{enumitem}
\usepackage{pifont}
\usepackage{fancyvrb}
\usepackage{footnote}
\makesavenoteenv{tabular}
\makesavenoteenv{table}

\usetikzlibrary{arrows,shapes,positioning,shadows,trees}

\usepackage[font=small]{caption}

\usepackage{stmaryrd}

\def\by{{\boldsymbol{y}}}
\def\bx{{\boldsymbol{x}}}

\newcommand{\mnote}[1]
{
}

\pgfplotscreateplotcyclelist{mark list*}{%
every mark/.append style={solid,fill=.!80!black},mark=*\\%
every mark/.append style={solid,fill=.!80!black},mark=square*\\%
every mark/.append style={solid,fill=.!80!black},mark=triangle*\\%
every mark/.append style={solid,fill=.!80!black},mark=halfsquare*\\%
every mark/.append style={solid,fill=.!80!black},mark=pentagon*\\%
every mark/.append style={solid,fill=.!80!black},mark=halfcircle*\\%
every mark/.append style={solid,fill=.!80!black,rotate=180},mark=halfdiamond*\\%
every mark/.append style={solid,fill=.!80!black!40},mark=otimes*\\%
every mark/.append style={solid,fill=.!80!black},mark=diamond*\\%
every mark/.append style={solid,fill=.!80!black},mark=halfsquare right*\\%
every mark/.append style={solid,fill=.!80!black},mark=halfsquare left*\\%
}
\pgfplotscreateplotcyclelist{color list}{%
red,blue,black,yellow,brown,teal,orange,violet,cyan,green!70!black,magenta,gray}
\pgfplotscreateplotcyclelist{linestyles}{solid,dashed,dotted}
\pgfplotscreateplotcyclelist{linestyles*}{solid,dashed,dotted,dashdotted,dashdotdotted}

\tikzset{every mark/.append style={scale=1.5}}

\pgfplotscreateplotcyclelist{mylist}{%
{blue,mark=*,solid},
{green!60!black,mark=triangle},
{brown!60!black,mark=diamond*},
{brown!60!black,mark options={fill=brown!40},mark=otimes*},
{yellow!60!black,mark=triangle*,mark options={fill=brown!40}},
{red,mark=square,}}

\DeclareMathOperator*{\argmin}{arg\,min}

\newcommand*{\Let}[2]{\State {#1} $\gets$ {#2}}

\enlargethispage{\baselineskip} 

\newcommand\Mark[1]{\textsuperscript#1}

\newcommand{\ignore}[1]{}

\aclfinalcopy 
\def\aclpaperid{693} 

\title{Cross-lingual Models of Word Embeddings: An Empirical Comparison}
\author{
  Shyam Upadhyay\Mark{1} ~ Manaal Faruqui\Mark{2} ~ Chris Dyer\Mark{2} ~ Dan Roth\Mark{1}\\
  \Mark{1} Department of Computer Science, University of Illinois, Urbana-Champaign, IL, USA\\
  \Mark{2} School of Computer Science, Carnegie Mellon University, Pittsburgh, PA, USA \\
       {\tt upadhya3@illinois.edu, mfaruqui@cs.cmu.edu} \\
       {\tt cdyer@cs.cmu.edu, danr@illinois.edu}
  }
\date{}

\begin{document}

\maketitle

\begin{abstract}
Despite interest in using cross-lingual knowledge to learn word
embeddings for various tasks, a systematic comparison of the possible
approaches is lacking in the literature. We perform an extensive
evaluation of four popular approaches of inducing cross-lingual
embeddings, each requiring a different form of supervision, on four
typologically different language pairs. Our evaluation setup spans
four different tasks, including intrinsic evaluation on mono-lingual
and cross-lingual similarity, and extrinsic evaluation on downstream
semantic and syntactic applications. We show that models which require
expensive cross-lingual knowledge almost always perform better, but
cheaply supervised models often prove competitive on certain tasks.


\end{abstract}

\section{Introduction}
Learning word vector representations using monolingual
distributional information is now a ubiquitous technique in
NLP. The quality of these word vectors can be significantly improved by
incorporating cross-lingual distributional information
\cite[\emph{inter
    alia}]{klementiev,zou:2013,vulic-moens:2013:EMNLP,mikolov2013exploiting,faruqui:2014,Hermann:2014:ACLphil,lauly:2014}, with
improvements observed both on monolingual
\cite{faruqui:2014,rastogi2015multiview} and cross-lingual tasks~\cite{guocross15,sogaard-EtAl:2015:ACL-IJCNLP,guocross16}.

Several models for inducing cross-lingual embeddings have been
proposed, each requiring a different form of cross-lingual supervision
-- some can use document-level alignments~\cite{vulic:2015}, others
need alignments at the sentence~\cite{Hermann:2014:ACLphil,gouws:2015}
or word level~\cite{faruqui:2014,gouws-sogaard:2015:NAACL-HLT}, while
some require both sentence and word
alignments~\cite{luong:2015}. However, a systematic comparison of
these models is missing from the literature, making it difficult to
analyze which approach is suitable for a particular NLP task. In this
paper, we fill this void by empirically comparing four cross-lingual
word embedding models each of which require different form of
alignment(s) as supervision, across several dimensions. To this end,
we train these models on four different language pairs, and evaluate
them on both monolingual and cross-lingual
tasks.\footnote{Instructions and code to reproduce the
  experiments available at
  \url{http://cogcomp.cs.illinois.edu/page/publication_view/794}}

First, we show that different models can be viewed as instances of a more
general framework for inducing cross-lingual word embeddings. Then, we
evaluate these models on both extrinsic and intrinsic tasks. Our
intrinsic evaluation assesses the quality of the vectors on
monolingual (\S\ref{sec:mono}) and cross-lingual (\S\ref{sec:dict})
word similarity tasks, while our extrinsic evaluation spans semantic
(cross-lingual document classification \S\ref{sec:doc}) and
syntactic tasks (cross-lingual dependency parsing \S\ref{sec:parsing}).

Our experiments show that word vectors trained using expensive
cross-lingual supervision (word alignments or sentence alignments)
perform the best on semantic tasks. On the other hand, for syntactic
tasks like cross-lingual dependency parsing, models requiring weaker
form of cross-lingual supervision (such as context agnostic
translation dictionary) are competitive to models requiring expensive
supervision. We also show qualitatively how the nature of
cross-lingual supervision used to train word vectors affects the
proximity of translation pairs across languages, and of words with
similar meaning in the same language in the vector-space.

\ignore{
Our experiments show that models which require expensive
cross-lingual supervision (word alignments or sentence alignments) are
superior on most tasks.  Nevertheless, their advantage over relatively
weakly supervised models varies.
 In fact, in tasks like cross-lingual
dependency parsing, models requiring weaker supervision perform
slightly better.

We posit that in downstream syntactic tasks like
dependency parsing, the form (word level, sentence level or both) of
the supervision is more crucial than its cost. Finally, we analyze the
embeddings qualitatively with the aim of revealing latent regularities
in the vector space which explain why some models do better than
others.
}


\section{Bilingual Embeddings}
\begin{figure}[tb]
  \centering
  \includegraphics[width=\columnwidth]{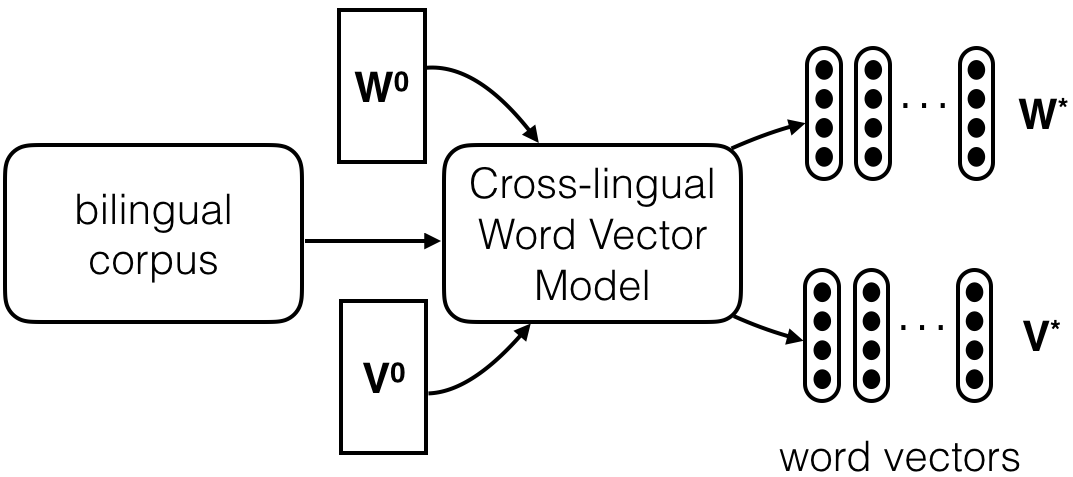}
  \begin{algorithm}[H]
    \footnotesize
    \caption{General Algorithm}
    \begin{algorithmic}[1]
      \State{Initialize $\mathbf{W} \leftarrow \mathbf{W}^0$,$\mathbf{V} \leftarrow \mathbf{V}^0$}
      \Let{$(\mathbf{W}^*,\mathbf{V}^*)$}{$\argmin \alpha A(\mathbf{W}) + \beta B(\mathbf{V}) + C(\mathbf{W},\mathbf{V})$}
    \end{algorithmic}
    \label{algo:general}
  \end{algorithm}
  \caption{({\bf Above}) A general schema for induction
    of cross-lingual word vector representations. The word
    vector model generates embeddings which incorporates
    distributional information {\em cross-lingually}. ({\bf Below}) A
    general algorithm for inducing bilingual word embeddings, where
    $\alpha,\beta,\mathbf{W}^0,\mathbf{V}^0$ are parameters and
    $A,B,C$ are suitably defined losses.}
  \label{fig:schema}
\end{figure}

A general schema for inducing bilingual embeddings is shown in
Figure~\ref{fig:schema}. Our comparison focuses on dense, fixed-length
distributed embeddings which are obtained using some form of
cross-lingual supervision.
We briefly describe the embedding induction procedure for each of
the selected bilingual word vector models, with the aim to provide a unified algorithmic perspective for all methods, and to facilitate better understanding and comparison. 
Our choice of models spans across different forms of supervision required for
inducing the embeddings, illustrated in Figure~\ref{fig:supervision}.

\paragraph{Notation.}
Let $W = \{ w_1, w_2, \ldots, w_{|W|}\}$ be the vocabulary of a
language $l_1$ with $|W|$ words, and $\mathbf{W} \in
\mathbb{R}^{|W| \times l}$ be the corresponding word embeddings of
length $l$. Let $V =\{ v_1, v_2, \ldots, v_{|V|}\}$ be the vocabulary
of another language $l_2$ with $|V|$ words, and $\mathbf{V} \in
\mathbb{R}^{|V| \times m}$ the corresponding word embeddings of
length $m$. 
We denote the word vector for a word $w$ by $\mathbf{w}$.
\ignore{
These are
not to be confused with a collection of monolingual word vectors
trained for different languages individually \cite{rfou}. Also,
approaches like CL-ESA~\cite{Sorg2012} which induce sparse, high
dimensional cross-lingual embeddings do not fall under the scope of
our comparison. We deliberately choose models which require different
levels of supervision, and are easy to train. Our choice was also
driven by availability of code, so that our results can be
reproduced.
}

\tikzset{MyStyle/.style={shape=rectangle,rounded corners,draw, align=center}}
\begin{figure*}
  \centering
  \begin{subfigure}[b]{0.2\textwidth}
    \centering
    \begin{tikzpicture}
      \node[MyStyle] at (0,0) (A) {\small I};
      \node[MyStyle] at (1,0) (B) {\small Love};
      \node[MyStyle] at (2,0) (C) {\small You};
      \node[MyStyle] at (0,1) (D) {\small Je};
      \node[MyStyle] at (1,1) (E) {\small t'};
      \node[MyStyle] at (2,1) (F) {\small aime};
      \draw[thick] (A) -- (D);
      \draw[thick] (B) -- (F);
      \draw[thick] (C) -- (E);
    \end{tikzpicture}
    \caption{{\bf BiSkip}}
    \label{fig:biskip}
  \end{subfigure}
  \begin{subfigure}[b]{0.2\textwidth}
    \centering
    \begin{tikzpicture}
      \node at (0,0) (A) {\small \underline{I Love You}};
      \node at (0,1) (D) {\small \underline{Je t' aime}};
      \draw[thick] (A) -- (D);
    \end{tikzpicture}
    \caption{{\bf BiCVM}}
    \label{fig:bicvm}
  \end{subfigure}
  \begin{subfigure}[b]{0.2\textwidth}
    \centering
    \begin{tikzpicture}
      \node[] at (0,0) (A) {\footnotesize (I, Je)};
      \node[] at (0,0.6) (C) {\footnotesize (Love, aime)};
      \node[] at (0,1.2) (E) {\footnotesize (You, t')};
    \end{tikzpicture}
    \caption{{\bf BiCCA}}
    \label{fig:bicca}
  \end{subfigure}
  \begin{subfigure}[b]{0.2\textwidth}
    \centering
    \begin{tikzpicture}
      \node[MyStyle] at (0,0) (A) [text width=3.5cm]{\footnotesize Hello! how are you? I Love You.};
      \node[MyStyle] at (0,1) (D) {\footnotesize Bonjour! Je t' aime.};
      \draw[thick](A) -- (D);
    \end{tikzpicture}
    \caption{{\bf BiVCD}}
    \label{fig:bivcd}
  \end{subfigure}
  \caption{Forms of supervision required by the four models compared in this paper. From left to right, the cost of the supervision required varies from expensive (BiSkip) to cheap (BiVCD). BiSkip requires a parallel corpus annotated with word alignments (Fig.~\ref{fig:biskip}), BiCVM requires a sentence-aligned corpus (Fig.~\ref{fig:bicvm}), BiCCA only requires a bilingual lexicon (Fig.~\ref{fig:bicca}) and BiVCD requires comparable documents (Fig.~\ref{fig:bivcd}).}
  \label{fig:supervision}
\end{figure*}
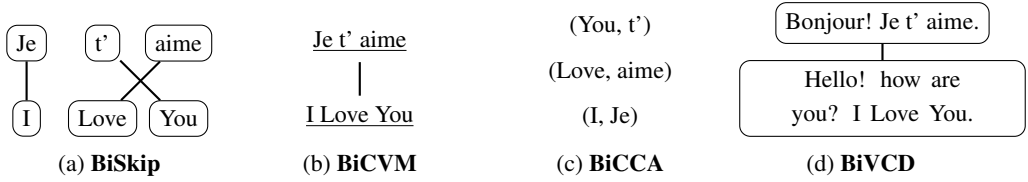


\subsection{Bilingual Skip-Gram Model (BiSkip)}
\label{sec:biskip}
\newcite{luong:2015} proposed Bilingual Skip-Gram, a simple extension
of the monolingual skip-gram model, which learns bilingual embeddings
by using a parallel corpus along with word alignments (both sentence
and word level alignments).

The learning objective is a simple extension of the skip-gram model,
where the context of a word is expanded to include bilingual links
obtained from word alignments, so that the model is trained to predict
words cross-lingually. In particular, given a word alignment link from
word $v \in V$ in language $l_2$ to $w \in W$ in language $l_1$, the model predicts the context words of
$w$ using $v$ and vice-versa. Formally, the cross
lingual part of the objective is,
\begin{equation}
D_{12}(\mathbf{W},\mathbf{V}) = - \sum_{(v,w) \in Q}\sum_{w_c \in \text{{\sc Nbr}}_1(w)}\log
P(w_c \mid v)
\end{equation}
where $\text{{\sc Nbr}}_1(w)$ is the context of $w$
in language $l_1$, $Q$ is the set of word alignments, and
$P(w_c \mid v) \propto
\exp({\mathbf{w}_c^T\mathbf{v}})$.  Another similar term $D_{21}$
models the objective for $v$ and $\text{{\sc
    Nbr}}_2(v)$.  The objective can be cast into Algorithm
\ref{algo:general} as,
\begin{equation}
  C(\mathbf{W},\mathbf{V}) = D_{12}(\mathbf{W},\mathbf{V}) + D_{21}(\mathbf{W},\mathbf{V})
\end{equation}
\begin{equation}
  A(\mathbf{W}) = - \sum_{w \in W}\sum_{w_c \in \text{{\sc Nbr}}_1(w)}\log P(w_c \mid w)
\end{equation}
\begin{equation}
  B(\mathbf{V}) = - \sum_{v \in V}\sum_{v_c \in \text{{\sc Nbr}}_2(v)}\log P(v_c \mid v)
\end{equation}
where $A(\mathbf{W})$ and $B(\mathbf{V})$ are the familiar skip-gram
formulation of the monolingual part of the objective. $\alpha$ and $\beta$ are chosen hyper-parameters which set the relative importance of the monolingual terms.

\subsection{Bilingual Compositional Model (BiCVM)}
\label{sec:bicvm}
\newcite{Hermann:2014:ACLphil} present a method that learns bilingual
word vectors from a sentence aligned corpus. Their model leverages the
fact that aligned sentences have equivalent meaning, thus their
sentence representations should be similar.

We denote two aligned sentences, $\vec{v} = \langle \bx_1, \ldots,
\rangle$ and $\vec{w} = \langle \by_1, \ldots \rangle$ , where $\bx_i \in
\mathbf{V}, \by_i \in \mathbf{W}$, are vectors corresponding to the
words in the sentences. Let functions $f:\vec{v} \rightarrow
\mathbb{R}^{n}$ and $g:\vec{w} \rightarrow \mathbb{R}^{n}$, map
sentences to their semantic representations in $\mathbb{R}^{n}$. BiCVM
generates word vectors by minimizing the squared $\ell_2$ norm between
the sentence representations of aligned sentences.  In order to
prevent the degeneracy arising from directly minimizing the $\ell_2$ norm,
they use a noise-contrastive large-margin update, with randomly drawn
sentence pairs $(\vec{v},\vec{w}^n)$ as negative samples. The loss for
the sentence pairs $(\vec{v},\vec{w})$ and $(\vec{v},\vec{w}^n)$ can
be written as,
\begin{equation}
  E(\vec{v}, \vec{w}, \vec{w}^{n}) = \mathrm{max}\left(\delta + \Delta
  E(\vec{v},\vec{w},\vec{w}^{n}), 0 \right)
\end{equation}
where,
\begin{equation}
E(\vec{v},\vec{w}) = \Vert f(\vec{v}) - g(\vec{w})\Vert^{2}
\end{equation}
and,
\begin{equation}
\Delta E(\vec{v},\vec{w},\vec{w}^{n}) = E(\vec{v}, \vec{w})- E(\vec{v},\vec{w}^{n})
\end{equation}
This can be cast into Algorithm \ref{algo:general} by,
\begin{equation}
  C(\mathbf{W},\mathbf{V}) = \sum_{\substack{\text{aligned }(\vec{v},
      \vec{w})\\\text{random }\vec{w}^{n} }} E(\vec{v}, \vec{w},
  \vec{w}^{n})
\end{equation}
\begin{equation}
\label{equ:reg}
\begin{array}{cc}
A(\mathbf{W}) =  \|\mathbf{W}\|^2 & B(\mathbf{V}) = \|\mathbf{V}\|^2
\end{array}
\end{equation}
with $A(\mathbf{W})$ and $B(\mathbf{V})$ being regularizers, with $\alpha=\beta$.

\subsection{Bilingual Correlation Based Embeddings (BiCCA)}
\label{sec:cca}
The BiCCA model, proposed by \newcite{faruqui:2014}, showed that when
(independently trained) monolingual vector matrices $\mathbf{W}, \mathbf{V}$ are projected
using CCA~\cite{hotellingCCA} to respect a translation lexicon, their
performance improves on word similarity and word analogy tasks.  They
first construct $\mathbf{W}' \subseteq \mathbf{W}, \mathbf{V}'
\subseteq \mathbf{V}$ such that $|\mathbf{W}'| = |\mathbf{V}'|$ and
the corresponding words ($w_i, v_i$) in the matrices are translations
of each other. The projection is then computed as:
\begin{equation}
\label{equ:cca}
\mathbf{P}_{W}, \mathbf{P}_{V} = \mathrm{CCA} (\mathbf{W}', \mathbf{V}')
\end{equation}
\begin{equation}
\label{equ:full-proj}
\begin{array}{cc}
\mathbf{W}^{*} =  \mathbf{W} \mathbf{P}_{W} & \mathbf{V}^{*} = \mathbf{V} \mathbf{P}_{V}
\end{array}
\end{equation}
where, $\mathbf{P}_{V} \in \mathbb{R}^{l \times d}, \mathbf{P}_{W} \in \mathbb{R}^{m \times d}$
are the projection matrices with $d \le \mathrm{min}(l, m)$ and the
$\mathbf{V}^{*} \in \mathbb{R}^{|V| \times d}, \mathbf{W}^{*} \in
\mathbb{R}^{|W| \times d}$ are the word vectors that have been
``enriched'' using bilingual knowledge.

The BiCCA objective can be viewed\footnote{described in Section 6.5
  of~\cite{hardoon2004canonical}} as the following instantiation of
Algorithm \ref{algo:general}:
\begin{equation}
  \mathbf{W}^0 = \mathbf{W}',\mathbf{V}^0 = \mathbf{V}'
\end{equation}
\begin{equation}
C(\mathbf{W},\mathbf{V}) = \|\mathbf{W}-\mathbf{V}\|^2 + \gamma
\left( \mathbf{V}^T \mathbf{W} \right)
\end{equation}
\begin{equation}
  \begin{array}{cc}
    A(\mathbf{W}) = \|\mathbf{W}\|^2-1 & B(\mathbf{V}) = \|\mathbf{V}\|^2-1
  \end{array}
\end{equation}
where $\mathbf{W}=\mathbf{W}^0 \mathbf{P}_W$ and $\mathbf{V}=\mathbf{V}^0 \mathbf{P}_V$, where we set $\alpha=\beta=\gamma=\infty$ to set hard constraints. 

\subsection{Bilingual Vectors from Comparable Data (BiVCD)}
\label{sec:comparable}
Another approach of inducing bilingual word vectors, which we refer to
as BiVCD, was proposed by \newcite{vulic:2015}. Their approach is
designed to use {\em comparable} corpus between the source and target
language pair to induce cross-lingual vectors.

Let $d_e$ and $d_f$ denote a pair of comparable documents with length
in words $p$ and $q$ respectively (assume $p>q$). BiVCD first merges these two
comparable documents into a single pseudo-bilingual document using a
deterministic strategy based on length ratio of two documents
$R=\lfloor \frac{p}{q} \rfloor$. Every $R^{th}$ word of the merged
pseudo-bilingual document is picked sequentially from $d_f$. Finally,
a skip-gram model is trained on the corpus of pseudo-bilingual
documents, to generate vectors for all words in $\mathbf{W}^{*} \cup
\mathbf{V}^{*}$. The vectors constituting $\mathbf{W}^{*}$ and
$\mathbf{V}^{*}$ can then be easily identified.

Instantiating BiVCD in the general algorithm is obvious: $C(\mathbf{W},\mathbf{V})$ assumes the familiar word2vec skip-gram
objective over the pseudo-bilingual document,
\begin{equation}
  C(\mathbf{W},\mathbf{V}) = - \sum_{s \in W\cup V}\sum_{t \in \text{{\sc Nbr}}(s)}\log P(t \mid s)
\end{equation}
where {\sc Nbr($s$)} is defined by the pseudo-bilingual
document and $P(t \mid s) \propto
\exp({\mathbf{t}^T\mathbf{s}})$. Note that $t,s \in W\cup V$.

Although BiVCD is designed to use comparable corpus, we provide it
with parallel data in our experiments (to ensure comparability), and
treat two aligned sentences as comparable.


\section{Data}
\label{sec:data}
We train cross-lingual embeddings for 4 language pairs: English-German
(en-de), English-French (en-fr), English-Swedish (en-sv) and
English-Chinese (en-zh). For en-de and en-sv we use the Europarl v7
parallel
corpus\footnote{\url{www.statmt.org/europarl/v7/{de,sv}-en.tgz}}~\cite{koehn2005europarl}.
For en-fr, we use Europarl combined with the news-commentary and
UN-corpus dataset from WMT
2015.\footnote{\url{www.statmt.org/wmt15/translation-task.html}} For
en-zh, we use the FBIS parallel corpus from the news domain
(LDC2003E14). We use the Stanford Chinese Segmenter~\cite{I05-3027} to
preprocess the en-zh parallel corpus. Corpus statistics for all
languages is shown in Table~\ref{tab:parallel}.
\begin{table}[!tb]
  \centering
  \begin{tabular}{llccc}
  \toprule
  $l_1$ & $l_2$ & \#sent & \#$l_1$-words & \#$l_2$-words\\
  \midrule
  \multirow{4}{*}{en} & de & 1.9 & 53 & 51\\
   & fr & 2.0 & 55 & 61\\
   & sv & 1.7 & 46 & 42\\
   & zh & 2.0 & 58 & 50\\
  \bottomrule
  \end{tabular}
  \caption{The size of parallel corpora (in millions) of different
    language pairs used for training cross-lingual word vectors.}
  \label{tab:parallel}
\end{table}

\section{Evaluation}
We measure the quality of the induced cross-lingual word embeddings in terms of
their performance, when used as features in the following tasks:
\begin{itemize}
\item monolingual word similarity for English
\item Cross-lingual dictionary induction
\item Cross-lingual document classification
\item Cross-lingual syntactic dependency parsing
\end{itemize}

The first two tasks intrinsically measure how much can monolingual and
cross-lingual similarity benefit from cross-lingual training. The last
two tasks measure the ability of cross-lingually trained vectors to
extrinsically facilitate model transfer across languages, for semantic
and syntactic applications respectively. These tasks have been used in
previous
works~\cite{klementiev,luong:2015,vulic-moens:2013:NAACL-HLT,guocross15}
for evaluating cross-lingual embeddings, but no comparison exists which
uses them in conjunction.

To ensure fair comparison, all models are trained with embeddings of
size 200. We provide all models with parallel corpora, irrespective of
their requirements. Whenever possible, we also report statistical
significance of our results.


\subsection{Parameter Selection}
We follow the {\bf BestAvg} parameter selection strategy from
\newcite{lu2015deep}: we selected the parameters for all models by tuning
on a set of values (described below) and picking the parameter setting
which did best on an average across all tasks.

\paragraph{BiSkip.}
All models were trained using a window size of 10 (tuned over
$\{5,10,20\}$), and 30 negative samples (tuned over
$\{10,20,30\}$). The cross-lingual weight was set to 4 (tuned over
$\{1,2,4,8\}$). The word alignments for training the
model (available at {\url{github.com/lmthang/bivec}}) were generated using
\texttt{fast_align}~\cite{dyer2013simple}. The number of training iterations was set to
5 (no tuning) and we set $\alpha=1$ and $\beta=1$ (no tuning).

\paragraph{BiCVM.}
We use the tool (available at {\url{github.com/karlmoritz/bicvm}}) released
by~\newcite{Hermann:2014:ACLphil} to train all embeddings. We train an
additive model (that is, $f(\vec{x})=g(\vec{x})=\sum_i x_i$) with
hinge loss margin set to 200 (no tuning), batch size of 50 (tuned over
${50,100,1000}$) and noise parameter of 10 (tuned over
$\{10,20,30\}$). All models are trained for 100 iterations (no
tuning).

\paragraph{BiCCA.}
First, monolingual word vectors are trained using the skip-gram
model\footnote{\url{code.google.com/p/word2vec}} with negative
sampling \cite{mikolov2013efficient} with window of size 5 (tuned over
$\{5,10,20\}$). To generate a cross-lingual dictionary, word alignments
are generated using cdec from the parallel corpus. Then, word pairs
$(a,b), a \in l_1, b \in l_2$ are selected such that $a$ is aligned to
$b$ the most number of times and vice versa. This way, we obtained dictionaries
of approximately 36k, 35k, 30k and 28k word pairs for en-de, en-fr,
en-sv and en-zh respectively.

The monolingual vectors are aligned using the above dictionaries with the
tool (available at {\url{github.com/mfaruqui/eacl14-cca}}) released
by \newcite{faruqui:2014} to generate the cross-lingual word
embeddings. We use $k=0.5$ as the number of canonical components
(tuned over $\{0.2,0.3,0.5,1.0\}$). Note that this results in a
embedding of size 100 after performing CCA.

\paragraph{BiVCD.}
We use word2vec's skip gram model for training our embeddings, with a
window size of 5 (tuned on $\{5,10,20,30\}$) and negative sampling
parameter set to 5 (tuned on $\{5,10,25\}$). Every pair of parallel
sentences is treated as a pair of comparable documents, and merging is
performed using the sentence length ratio strategy described
earlier.\footnote{We implemented the code for performing the merging as we
could not find a tool provided by the authors.}
\subsection{Monolingual Evaluation}
\label{sec:mono}
We first evaluate if the inclusion of cross-lingual knowledge
improves the quality of English embeddings.

\paragraph{Word Similarity.} Word similarity datasets 
contain word pairs which are assigned similarity ratings by
humans. The task evaluates how well the notion of word similarity
according to humans is emulated in the vector space. Evaluation is
based on the Spearman's rank correlation coefficient~\cite{spearman}
between human rankings and rankings produced by computing
cosine similarity between the vectors of two words.

We use the SimLex dataset for English~\cite{simlex} which contains 999
pairs of English words, with a balanced set of noun, adjective and
verb pairs. SimLex is claimed to capture word similarity exclusively
instead of WordSim-353~\cite{citeulike:379845} which captures both
word similarity and relatedness. We declare significant improvement if
$p < 0.1$ according to Steiger's method~\cite{steiger1980tests} for
calculating the statistical significant differences between two
dependent correlation coefficients.

Table~\ref{tab:wordsim} shows the performance of English embeddings
induced by all the models by training on different language pairs on
the SimLex word similarity task. The score obtained by monolingual
English embeddings trained on the respective English side of each
language is shown in column marked Mono.  In all cases (except BiCCA
on en-sv), the bilingually trained vectors achieve better scores than
the mono-lingually trained vectors.

Overall, across all language pairs, BiCVM is the best performing model
in terms of Spearman's correlation, but its improvement over BiSkip
and BiVCD is often insignificant. It is notable that 2 of the 3 top
performing models, BiCVM and BiVCD, need sentence aligned and
document-aligned corpus only, which are easier to obtain than parallel
data with word alignments required by BiSkip.


\begin{table}[!tb]
  \centering
  \begin{tabular}{l@{\,\,}c@{\,\,}@{\,\,}c@{\,\,}@{\,\,}c@{\,\,}@{\,\,}c@{\,\,}@{\,\,}c@{\,\,}}
  \toprule
  pair & Mono & BiSkip & BiCVM & BiCCA & BiVCD \\
  \midrule
  en-de & 0.29 & \underline{0.34} &  {\bf 0.37} & 0.30 & 0.32 \\
  en-fr & 0.30 & \underline{0.35} & {\bf 0.39} & 0.31 & 0.36 \\
  en-sv & 0.28 & \underline{0.32} & {\bf 0.34} & 0.27 & \underline{0.32} \\
  en-zh & 0.28 & \underline{0.34} & \underline{\bf 0.39} & 0.30 & 0.31 \\
  \midrule
  avg. & 0.29 & 0.34 & {\bf 0.37} & 0.30 & 0.33 \\
  \bottomrule
  \end{tabular}
  \caption{Word similarity score measured in Spearman's correlation
    ratio for English on SimLex-999. The best score for each language pair
    is shown in {\bf bold}. Scores which are significantly better (per
    Steiger's Method with $p<0.1$) than the next lower score are
    \underline{underlined}. For example, for en-zh, BiCVM is
    significantly better than BiSkip, which in turn is significantly
    better than BiVCD.}
  \label{tab:wordsim}
\end{table}
\paragraph{\textsc{Qvec}.} \newcite{qvec} proposed an intrinsic evaluation
metric for estimating the quality of English word vectors. The score
produced by \textsc{Qvec} measures how well a given set of word
vectors is able to quantify linguistic properties of words, with
higher being better. The metric is shown to have strong correlation
with performance on downstream semantic applications. As it can be
currently only used for English, we use it to evaluate the English
vectors obtained using cross-lingual training of different
models. Table~\ref{tab:qvec} shows that on average across language
pairs, BiSkip achieves the best score, followed by Mono (mono-lingually
trained English vectors), BiVCD and BiCCA. A possible explanation for
why Mono scores are better than those obtained by some of the
cross-lingual models is that {\sc Qvec} measures monolingual semantic
content based on a linguistic oracle made for English. Cross-lingual
training might affect these semantic properties
arbitrarily.

Interestingly, BiCVM which was the best model according
to SimLex, ranks last according to \textsc{Qvec}. The fact
that the best models according to \textsc{Qvec} and word similarities
are different reinforces observations made in previous work that
performance on word similarity tasks alone does not reflect
quantification of linguistic properties of words
\cite{qvec,schnabel-2015}.

\begin{table}[!tb]
  \centering
  \begin{tabular}{l@{\,\,}c@{\,\,}@{\,\,}c@{\,\,}@{\,\,}c@{\,\,}@{\,\,}c@{\,\,}@{\,\,}c@{\,\,}}
  \toprule
  pair & Mono & BiSkip & BiCVM & BiCCA & BiVCD \\
  \midrule
  en-de & 0.39 & {\bf 0.40} & 0.31 & 0.33 & 0.37\\
  en-fr & 0.39 & {\bf 0.40} & 0.31 & 0.33 & 0.38\\
  en-sv & 0.39 & {\bf 0.39} & 0.31 & 0.32 & 0.37\\
  en-zh & 0.40 & {\bf 0.40} & 0.32 & 0.33 & 0.38\\
  \midrule
  avg. & 0.39 & {\bf 0.40} & 0.31 & 0.33 & 0.38\\
  \bottomrule
  \end{tabular}
  \caption{Intrinsic evaluation of English word vectors measured in terms of
\textsc{Qvec} score across models. Best scores for each language pair is shown in {\bf bold}.}
  \label{tab:qvec}
\end{table}


\subsection{Cross-lingual Dictionary Induction}
\label{sec:dict}
The task of cross-lingual dictionary induction~\cite{vulic-moens:2013:NAACL-HLT,gouws:2015,mikolov2013exploiting} judges how good
cross-lingual embeddings are at detecting word pairs that are semantically
similar across languages. We follow the setup of
\newcite{vulic-moens:2013:NAACL-HLT}, but instead of manually creating
a gold cross-lingual dictionary, we derived our gold dictionaries using the Open
Multilingual WordNet data released by
\newcite{bond-foster:2013:ACL2013}. The data includes synset
alignments across 26 languages with over 90\% accuracy. First, we
prune out words from each synset whose frequency count is less than
$1000$ in the vocabulary of the training data from
\S\ref{sec:data}. Then, for each pair of aligned synsets
$s_1=\{k_1,k_2,\cdots\}$ $s_2=\{g_1,g_2,\cdots\}$, we include all
elements from the set $\{(k,g) \mid k \in s_1, g \in s_2\}$ into the
gold dictionary, where $k$ and $g$ are the lemmas. Using this approach
we generated dictionaries of sizes 1.5k, 1.4k, 1.0k and 1.6k pairs for
en-fr, en-de, en-sv and en-zh respectively.
\begin{table}[!tb]
  \centering
  \begin{tabular}{llcccc}
  \toprule
  $l_1$ & $l_2$ & BiSkip & BiCVM & BiCCA & BiVCD \\
  \midrule
  \multirow{4}{*}{en} & de & \textbf{79.7} & 74.5 & 72.4 & 62.5\\
   & fr & \textbf{78.9} & 72.9 & 70.1 & 68.8\\
   & sv & \textbf{77.1} & 76.7 & 74.2 & 56.9\\
   & zh & \textbf{69.4} & 66.0 & 59.6 & 53.2\\
 \midrule
  \multicolumn{2}{c}{avg.} & \textbf{76.3} & 72.5 & 69.1 & 60.4\\
  \bottomrule
  \end{tabular}
  \caption{Cross-lingual dictionary induction results (top-10
    accuracy). The same trend was also observed across models when computing MRR
    (mean reciprocal rank).}
  \label{tab:dict}
\end{table}

We report top-10 accuracy, which is the fraction of the entries
$(e,f)$ in the gold dictionary, for which $f$ belongs to the list of
top-10 neighbors of the word vector of $e$, according to the induced
cross-lingual embeddings.
From the results (Table \ref{tab:dict}), it can be seen that for
dictionary induction, the performance improves with the quality of
supervision. As we move from cheaply supervised methods (eg. BiVCD) to
more expensive supervision (eg. BiSkip), the accuracy improves. This
suggests that for cross lingual similarity tasks, the more expensive
the cross-lingual knowledge available, the better. Models using weak
supervision like BiVCD perform poorly in comparison to models like
BiSkip and BiCVM, with performance gaps upwards of 10 pts on an
average.


\subsection{Cross-lingual Document Classification}
\label{sec:doc}
\begin{table}[!tb]
  \centering
  \begin{tabular}{llcccc}
  \toprule
  $l_1$ & $l_2$ & BiSkip & BiCVM & BiCCA & BiVCD \\
  \midrule
  \multirow{4}{*}{en} & de & {\bf 85.2} & \underline{85.0} & 79.1 & 79.9\\
   & fr & \underline{\bf 77.7} & 71.7 & 70.7 & 72.0\\
   & sv & \underline{\bf 72.3} & \underline{69.1} & \underline{65.3} & 59.9\\
   & zh & {\bf 75.5} & {73.6} & 69.4 & \underline{73.0}\\
  \midrule
  de & \multirow{4}{*}{en} & {\bf 74.9} & \underline{71.1} & 64.9 & \underline{74.1}\\
  fr & & \underline{\bf 80.4} & 73.7 & \underline{75.5} & \underline{77.6}\\
  sv & & \underline{73.4} & 67.7 & 67.0 & \underline{\bf 78.2}\\
  zh & & {\bf 81.1} & 76.4 & 77.3 & \underline{80.9}\\
  \midrule
  \multicolumn{2}{c}{avg.} & {\bf 77.6} & 73.5 & 71.2 & 74.5\\
  \bottomrule
  \end{tabular}
  \caption{Cross-lingual document classification accuracy when trained
    on language $l_1$, and evaluated on language $l_2$. The best score
    for each language is shown in {\bf bold}. Scores which are
    significantly better (per McNemar's Test with $p<0.05$) than the
    next lower score are \underline{underlined}. For example, for
    sv$\rightarrow$en, BiVCD is significantly better than BiSkip,
    which in turn is significantly better than BiCVM.}
  \label{tab:doc}
\end{table}

We follow the cross-lingual document classification (CLDC) setup of
\newcite{klementiev}, but extend it to cover all of our language
pairs. We use the RCV2 Reuters multilingual corpus\footnote{\url{http://trec.nist.gov/data/reuters/reuters.html}} for our
experiments.
In this task, for a language pair ($l_1, l_2$), a document classifier
is trained using the document representations derived from word
embeddings in language $l_1$, and then the trained model is tested on
documents from language $l_2$ (and vice-versa).  By using supervised
training data in one language and evaluating without further
supervision in another, CLDC assesses whether the learned cross-lingual
representations are semantically coherent across multiple languages.

All embeddings are learned on the data described in \S\ref{sec:data},
and we only use the RCV2 data to learn document classification
models. Following previous work, we compute document representation by
taking the tf-idf weighted average of vectors of the words present in
it.\footnote{tf-idf~\cite{Salton:1988:TAA:54259.54260} was computed
  using all documents for that language in RCV2.}
A multi-class classifier is trained using an averaged
perceptron~\cite{freund1999large} for 10 iterations, using the
document vectors of language $l_1$ as features\footnote{We use the
implementation of \newcite{klementiev}.}. 
Majority baselines for $en \rightarrow l_2$ and $l_1 \rightarrow en$
are $49.7\%$ and $46.7\%$ respectively, for all
languages. Table~\ref{tab:doc} shows the performance of different
models across different language pairs. We computed confidence values
using the McNemar test~\cite{mcnemar1947note} and declare significant
improvement if $p < 0.05$.

Table~\ref{tab:doc} shows that in almost all cases,
BiSkip performs significantly better than the remaining models. For
transferring semantic knowledge across languages via embeddings,
sentence and word level alignment proves superior to sentence or word
level alignment alone. This observation is consistent with the trend
in cross-lingual dictionary induction, where too the most expensive
form of supervision performed the best.

\subsection{Cross-lingual Dependency Parsing}
\label{sec:parsing}
Using cross lingual similarity for direct-transfer of dependency
parsers was first shown in \newcite{tackstrom2012cross}. The idea behind direct-transfer is
to train a dependency parsing model using embeddings for language
$l_1$ and then test the trained model on language $l_2$, replacing
embeddings for language $l_1$ with those of $l_2$. The transfer relies
on coherence of the embeddings across languages arising from the cross
lingual training. For our experiments, we use the cross lingual
transfer setup of \newcite{guocross15}.\footnote{
  \url{github.com/jiangfeng1124/acl15-clnndep}} Their framework trains
a transition-based dependency parser using nonlinear activation
function, with the source-side embeddings as lexical features. These
embeddings can be replaced by target-side embeddings at test time.

All models are trained for 5000 iterations with fixed word embeddings
during training. Since our goal is to determine the utility of word
embeddings in dependency parsing, we turn off other features that can
capture distributional information like brown clusters, which were
originally used in \newcite{guocross15}. We use the universal
dependency treebank~\cite{UnivDep} version-2.0 for our evaluation. For
Chinese, we use the treebank released as part of the CoNLL-X shared
task~\cite{buchholz2006conll}.

We first evaluate how useful the word embeddings are in cross-lingual
model transfer of dependency parsers (Table~\ref{tab:parsing}). On an
average, BiCCA does better than other models. BiSkip is a close
second, with an average performance gap of less than 1 point.  BiSkip
outperforms BiCVM on German and French (over 2 point improvement),
owing to word alignment information BiSkip's model uses during
training. It is not surprising that English-Chinese transfer scores
are low, due to the significant difference in syntactic structure of
the two languages. Surprisingly, unlike the semantic tasks considered
earlier, the models with expensive supervision requirements like
BiSkip and BiCVM could not outperform a cheaply supervised BiCCA.


\begin{table}[!tb]
  \centering
  \begin{tabular}{llcccc}
  \toprule
  $l_1$ & $l_2$ & BiSkip & BiCVM & BiCCA & BiVCD \\
  \midrule
  \multirow{4}{*}{en} & de & 49.8 & 47.5 & {\bf 51.3} & 49.0\\
   & fr & 65.8 & 63.2 & {\bf 65.9} & 60.7\\
   & sv & 56.9 & 56.7 & {\bf 59.4} & 54.6\\
  & zh & {\bf 6.4} & 6.1 & {\bf 6.4} & 6.0\\
  \midrule
  de & \multirow{4}{*}{en} & 49.7 & 45.0 & {\bf 50.3} & 43.6\\
  fr &  & 53.3 & 50.6 & {\bf 54.2} & 49.5\\
  sv &  & 48.2 & 49.0 & {\bf 49.9} & 44.6\\
  zh &  & {\bf 0.17} & 0.12 & {\bf 0.17} & 0.15\\
  \midrule
  \multicolumn{2}{c}{avg.} & 41.3 & 39.8 & {\bf 42.2} & 38.5\\
  \bottomrule
  \end{tabular}
  \caption{Labeled attachment score (LAS) for cross-lingual dependency parsing
  when trained on language $l_1$, and evaluated on language $l_2$. The best score
    for each language is shown in {\bf bold}.}
  \label{tab:parsing}
\end{table}

We also evaluate whether using cross-lingually trained vectors for
learning dependency parsers is better than using mono-lingually trained
vectors in Table~\ref{tab:direct}. We compare against parsing models
trained using mono-lingually trained word vectors (column marked Mono
in Table~\ref{tab:direct}).  These vectors are the same used as input
to the BiCCA model. All other settings remain the same. On an average
across language pairs, improvement over the monolingual embeddings was
obtained with the BiSkip and BiCCA models, while BiCVM and BiVCD
consistently performed worse. 
A possible reason for this is that BiCVM and BiVCD operate on sentence
level contexts to learn the embeddings, which only captures the
semantic meaning of the sentences and ignores the internal syntactic
structure. As a result, embedding trained using BiCVM and BiVCD are
not informative for syntactic tasks. On the other hand, BiSkip and
BiCCA both utilize the word alignment information to train their
embeddings and thus do better in capturing some notion of syntax.

\begin{table}[!tb]
  \centering
  \begin{tabular}{l@{\,\,}c@{\,\,}@{\,\,}c@{\,\,}@{\,\,}c@{\,\,}@{\,\,}c@{\,\,}@{\,\,}c@{\,\,}}
  \toprule
  $l$ & Mono & BiSkip & BiCVM & BiCCA & BiVCD \\
  \midrule
  de & 71.1 & {\bf 72.0} & 60.4 & {\bf 71.4} & 58.9\\
  fr & 78.9 & {\bf 80.4} & 73.7 & {\bf 80.2} & 69.5\\
  sv & 75.5 & {\bf 78.2} & 70.5 & {\bf 79.0} & 64.5\\
  zh & 73.8 & 73.1 & 65.8 & 71.7 & 67.0\\
  \midrule
  avg. & 74.8 & {\bf 75.9} & 67.6 & {\bf 75.6} & 66.8\\
  \bottomrule
  \end{tabular}
  \caption{Labeled attachment score (LAS) for dependency parsing when
    trained and tested on language $l$. Mono refers to parser trained
    with mono-lingually induced embeddings. Scores in {\bf bold} are
    better than the Mono scores for each language, showing improvement
    from cross-lingual training.}
  \label{tab:direct}
\end{table}




\section{Qualitative Analysis}
\label{sec:qual}
\begin{figure*}[ht]
  \begin{subfigure}[b]{0.5\linewidth}
    \centering
    \includegraphics[width=1.0\linewidth]{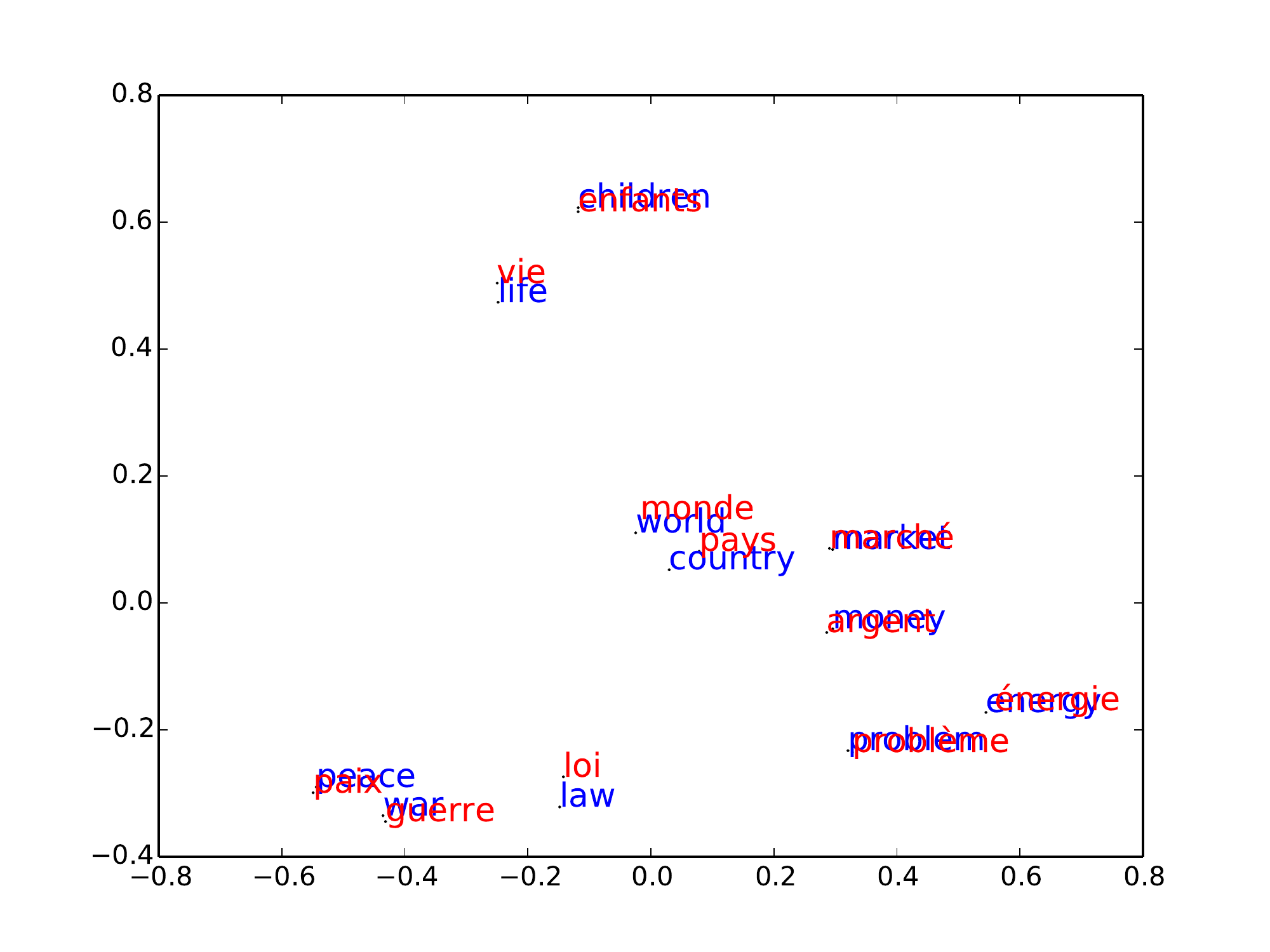}
    \caption{BiSkip}
    \vspace{2ex}
  \end{subfigure}
  \begin{subfigure}[b]{0.5\linewidth}
    \centering
    \includegraphics[width=1.0\linewidth]{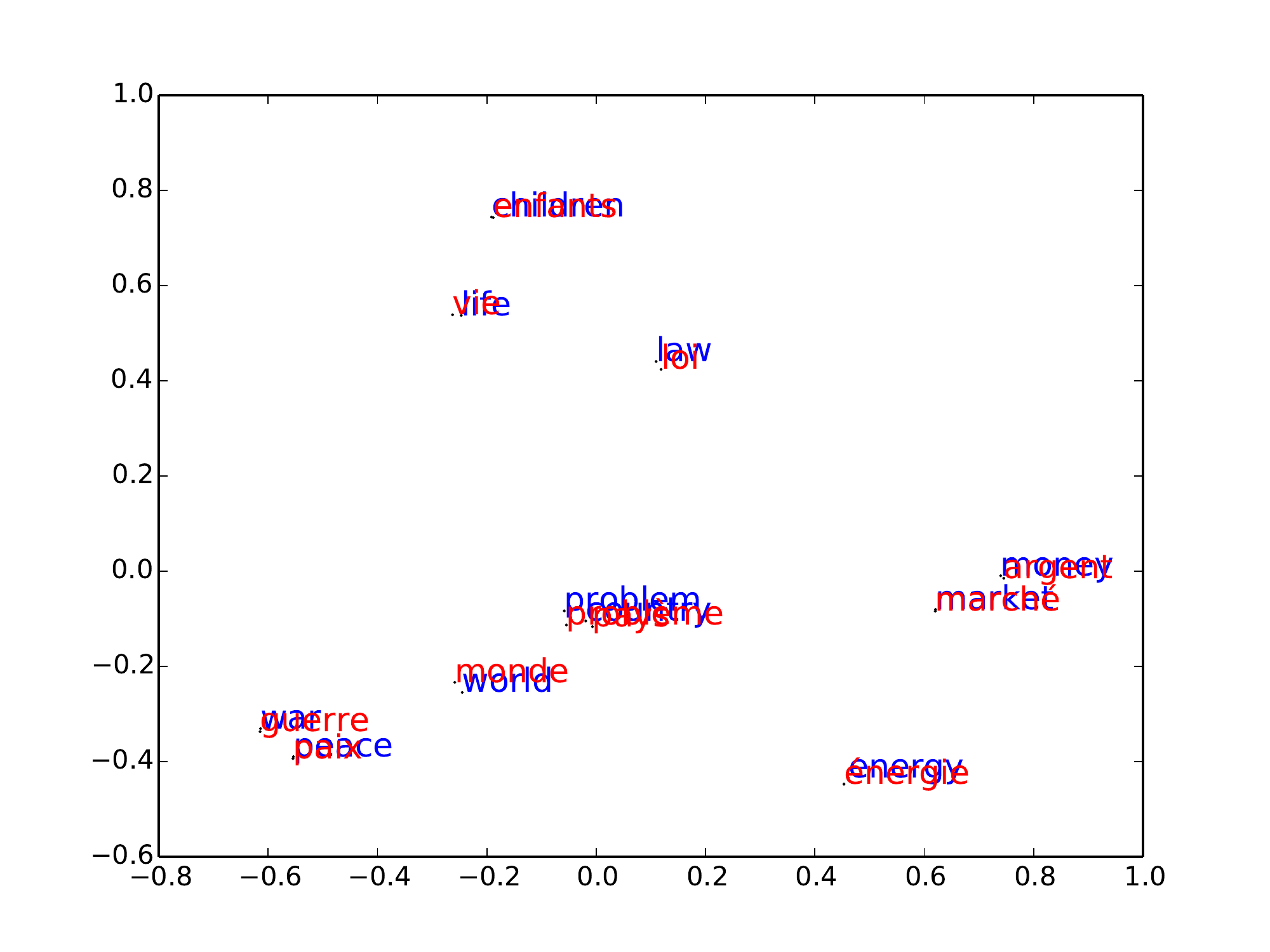}
    \caption{BiCVM}
    \vspace{2ex}
  \end{subfigure}
  \begin{subfigure}[b]{0.5\linewidth}
    \centering
    \includegraphics[width=1.0\linewidth]{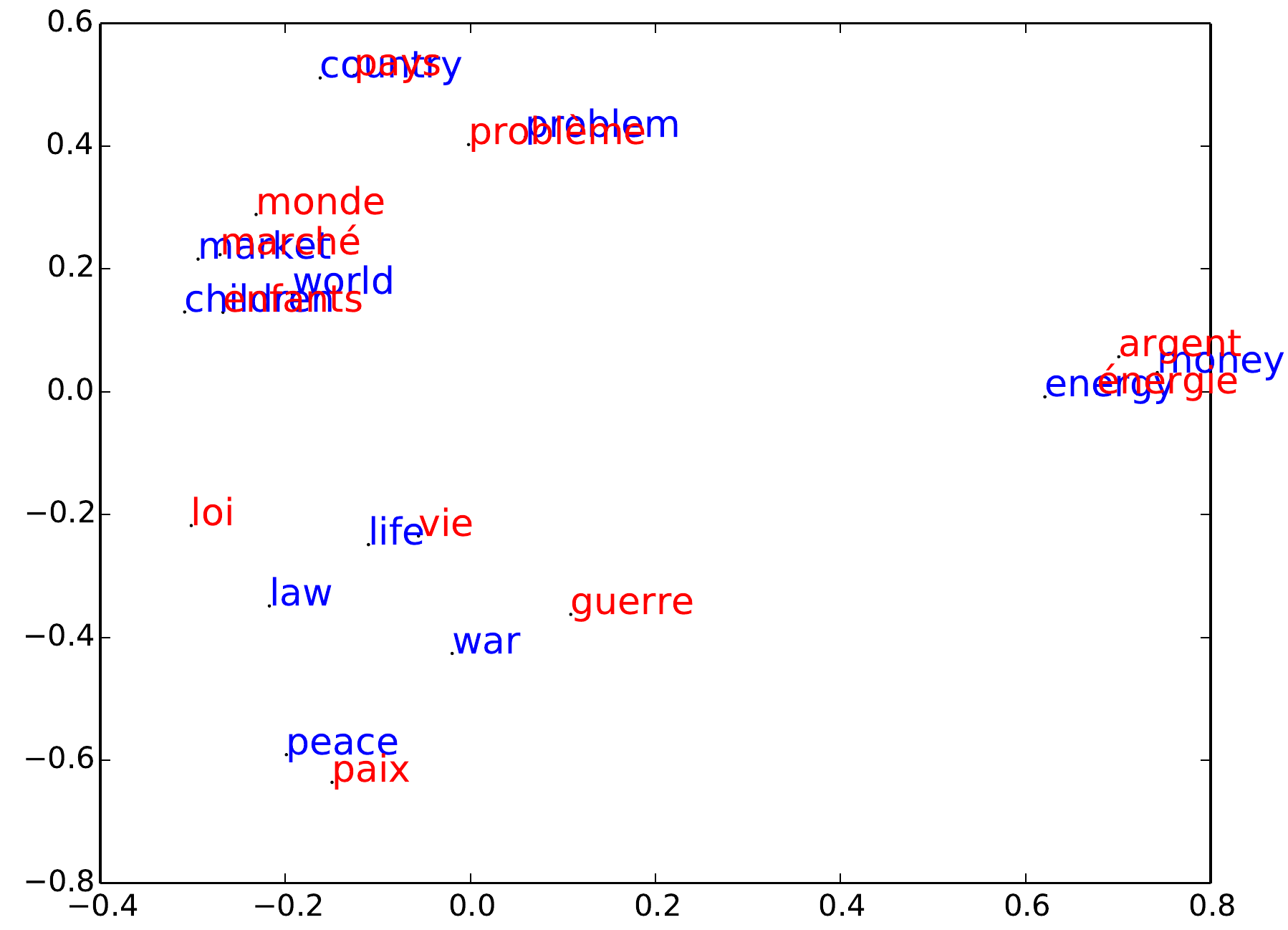}
    \caption{BiCCA}
    \vspace{2ex}
  \end{subfigure}
  \begin{subfigure}[b]{0.5\linewidth}
    \centering
    \includegraphics[width=1.0\linewidth]{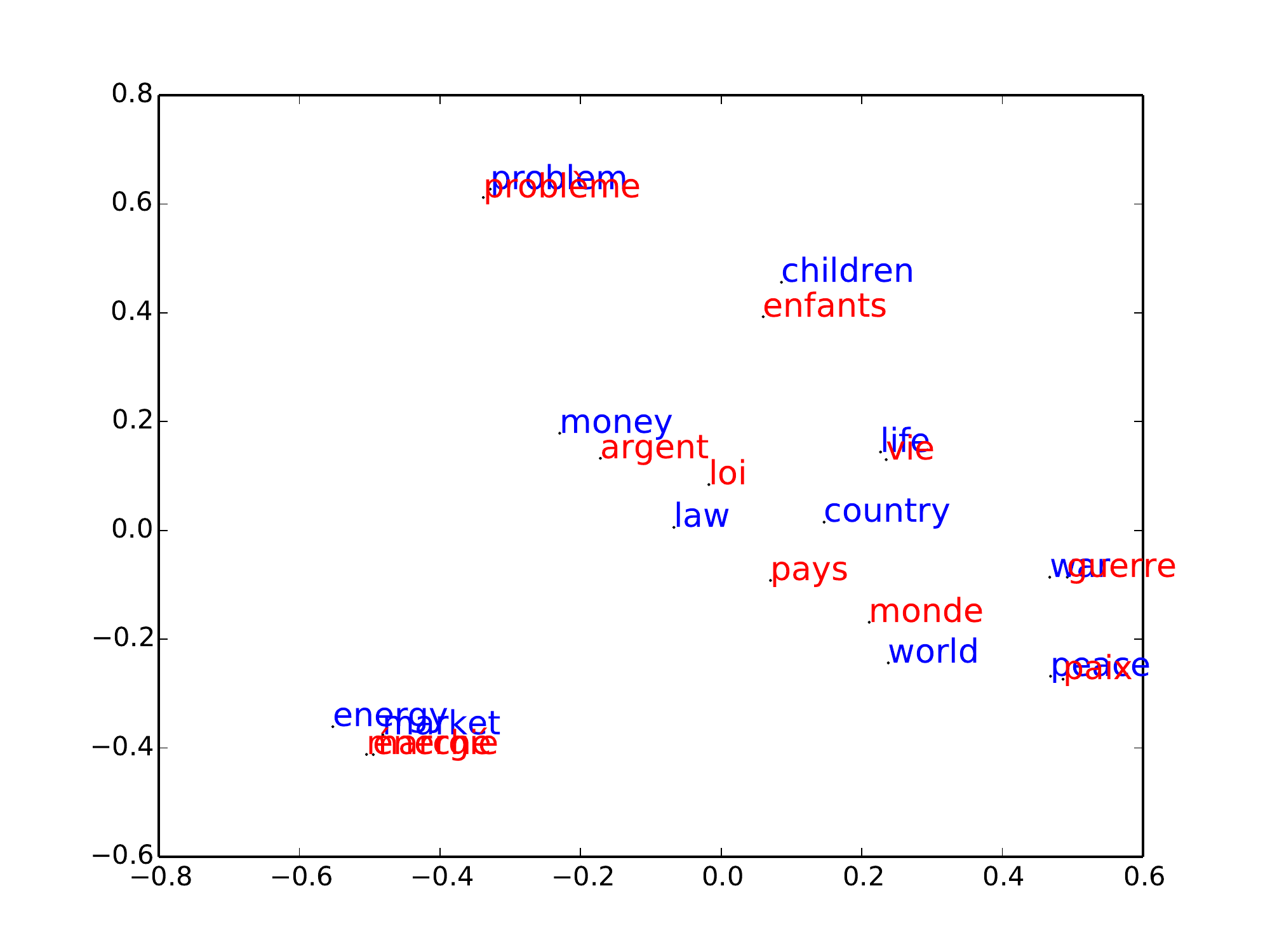}
    \caption{BiVCD}
    \vspace{2ex}
  \end{subfigure}
\caption{PCA projection of word embeddings of some frequent
  words present in English-French corpus. English and French words are
 shown in blue and red respectively.}
\label{fig:plots}
\end{figure*}

Figure~\ref{fig:plots} shows the PCA projection of some of the most frequent
words in the English-French corpus. It is clear that BiSkip and BiCVM produce
cross-lingual vectors which are the most comparable, the English and French
words which are translations of each other are represented by almost the same
point in the vector-space. In BiCCA and BiVCD the translated words are more
distant than BiSkip and BiCVM. This is not surprising because BiSkip and BiCVM
require more expensive supervision at the sentence level in contrast to the
other two models.

An interesting observation is that BiCCA and BiVCD are better at
separating antonyms. The words \textit{peace} and \textit{war}, (and
their French translations \textit{paix} and \textit{guerre}) are well
separated in BiCCA and BiVCD. However, in BiSkip and BiCVM these pairs
are very close together. This can be attributed to the fact that
BiSkip and BiCVM are trained on parallel sentences, and if two
antonyms are present in the same sentence in English, they will also
be present together in its French translation. However, BiCCA uses
bilingual dictionary and BiVCD use comparable sentence context, which
helps in pulling apart the synonyms and antonyms.



\section{Discussion}
The goal of this paper was to formulate the task of learning
cross-lingual word vector representations in a unified framework, and
conduct experiments to compare the performance of existing models in a
unbiased manner. We chose existing cross-lingual word vector models
that can be trained on two languages at a given time. In recent work,
\newcite{ammar2016massively} train multilingual word vectors using
more than two languages; our comparison does not cover this setting.
It is also worth noting that we compare here different cross-lingual
word embeddings, which are not to be confused with a collection of
monolingual word embeddings trained for different languages
individually~\cite{rfou}.

The paper does not cover {\em all} approaches that generate
cross-lingual word embeddings. Some methods do not have publicly
available code~\cite{coulmance-EtAl:2015,zou:2013}; for others, like
BilBOWA~\cite{gouws:2015}, we identified problems in the available
code, which caused it to consistently produced results that are
inferior even to mono-lingually trained vectors.\footnote{We contacted
the authors of the papers and were unable to resolve the issues in
the toolkit.}  However, the models that we included for comparison
in our survey are representative of other cross-lingual models in
terms of the form of cross-lingual supervision required by them. For
example, BilBOWA~\cite{gouws:2015} and cross-lingual Auto-encoder
\cite{lauly:2014} are similar to BiCVM in this respect. Multi-view
CCA~\cite{rastogi2015multiview} and deep CCA~\cite{lu2015deep} can be
viewed as extensions of BiCCA. Our choice of models was motivated to
compare different forms of supervision, and therefore, adding these
models, would not provide additional insight.

\section{Conclusion}

We presented the first systematic comparative evaluation of
cross-lingual embedding methods on several downstream NLP tasks, both
intrinsic and extrinsic. We provided a unified representation for all
approaches, showing them as instances of a general algorithm. Our
choice of methods spans a diverse range of approaches, in that each
requires a different form of supervision.

Our experiments reveal interesting trends. When evaluating on
intrinsic tasks such as monolingual word similarity, models relying on
cheaper forms of supervision (such as BiVCD) perform almost on par
with models requiring expensive supervision. On the other hand, for
cross-lingual semantic tasks, like cross-lingual document classification and dictionary induction, the
model with the most informative supervision performs best overall. In
contrast, for the syntactic task of dependency parsing, models that
are supervised at a word alignment level perform slightly
better. Overall this suggests that semantic tasks can benefit more from
richer cross-lingual supervision, as compared to syntactic tasks.

\subsection*{Acknowledgement}
\begin{small}
  This material is based on research sponsored by DARPA under
  agreement number FA8750-13-2-0008 and Contract
  HR0011-15-2-0025. Approved for Public Release, Distribution
  Unlimited. The views expressed are those of the authors and do not
  reflect the official policy or position of the Department of Defense
  or the U.S. Government.
\end{small}
\bibliography{references}
\bibliographystyle{acl2016}

\end{document}